\documentclass[sigconf]{acmart}
\AtBeginDocument{
 }

\copyrightyear{2023}
\acmYear{2023}
\setcopyright{rightsretained}
\acmConference[MM '23]{Proceedings of the 31st ACM International Conference on Multimedia}{October 29-November 3, 2023}{Ottawa, ON, Canada}
\acmBooktitle{Proceedings of the 31st ACM International Conference on Multimedia (MM '23), October 29-November 3, 2023, Ottawa, ON, Canada}
\acmDOI{10.1145/3581783.3612389}
\acmISBN{979-8-4007-0108-5/23/10}

\makeatletter
\gdef\@copyrightpermission{
  \begin{minipage}{0.3\columnwidth}
   \href{https://creativecommons.org/licenses/by/4.0/}{\includegraphics[width=0.90\textwidth]{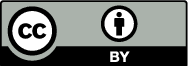}}
  \end{minipage}\hfill
  \begin{minipage}{0.7\columnwidth}
   \href{https://creativecommons.org/licenses/by/4.0/}{This work is licensed under a Creative Commons Attribution International 4.0 License.}
  \end{minipage}
  \vspace{5pt}
}
\makeatother

\usepackage{algorithm}
\usepackage[noend]{algpseudocode}
\usepackage{enumitem}
\usepackage{placeins}

\usepackage{graphicx,subcaption}
\usepackage{flushend} 
\newlist{noitemize}{itemize}{1}

\newcommand{\lxadd}[1]{\textcolor{blue}{#1}}
\newcommand{\ignore}[1]{}

\begin{document}

\title{Improving Zero-shot Visual Question Answering via Large Language Models with Reasoning Question Prompts}

\author{Yunshi Lan}
\affiliation{%
  \institution{East China Normal University}
  \city{Shanghai}
  \country{China}}
\email{yslan@dase.ecnu.edu.cn}

\author{Xiang Li}
\affiliation{%
  \institution{East China Normal University}
  \city{Shanghai}
  \country{China}}
\email{xiang.li@stu.ecnu.edu.cn}

\author{Xin Liu}
\affiliation{%
  \institution{East China Normal University}
  \city{Shanghai}
  \country{China}}
\email{xin.liu.0726@stu.ecnu.edu.cn}

\author{Yang Li}
\affiliation{%
  \institution{Alibaba Group}
  \city{Beijing}
  \country{China}}
\email{ly200170@alibaba-inc.com}

\author{Wei Qin}
\affiliation{%
  \institution{Hefei University of Technology}
  \city{Hefei}
  \country{China}}
\email{qinwei.hfut@gmail.com}

\author{Weining Qian}
\affiliation{%
  \institution{East China Normal University}
  \city{Shanghai}
  \country{China}}
\email{wnqian@dase.ecnu.edu.cn}
\renewcommand{\shortauthors}{Yunshi Lan et al.}

\begin{abstract}
Zero-shot Visual Question Answering (VQA) is a prominent vision-language task that examines both the visual and textual understanding capability of systems in the absence of training data. Recently, by converting the images into captions, information across multi-modalities is bridged and Large Language Models (LLMs) can apply their strong zero-shot generalization capability to unseen questions.
To design ideal prompts for solving VQA via LLMs, several studies have explored different strategies to select or generate question-answer pairs as the exemplar prompts, which guide LLMs to answer the current questions effectively.
However, they totally ignore the role of question prompts.
The original questions in VQA tasks usually encounter ellipses and ambiguity which require intermediate reasoning.
To this end, we present Reasoning Question Prompts for VQA tasks, which can further activate the potential of LLMs in zero-shot scenarios. Specifically, for each question, we first generate self-contained questions as reasoning question prompts via an unsupervised question edition module considering sentence fluency, semantic integrity and syntactic invariance.
Each reasoning question prompt clearly indicates the intent of the original question.
This results in a set of candidate answers.
Then, the candidate answers associated with their confidence scores acting as answer heuristics are fed into LLMs and produce the final answer.
We evaluate reasoning question prompts on three VQA challenges, experimental results demonstrate that they can significantly improve the results of LLMs on zero-shot setting and outperform existing state-of-the-art zero-shot methods on three out of four data sets.
Our source code is publicly released at \url{https://github.com/ECNU-DASE-NLP/RQP}.
\end{abstract}




\maketitle

\section{Introduction}
\label{sec:intro}

\begin{figure}[t!] 
    \centering
    \vspace{0.5cm}
    \includegraphics[width=0.45\textwidth]{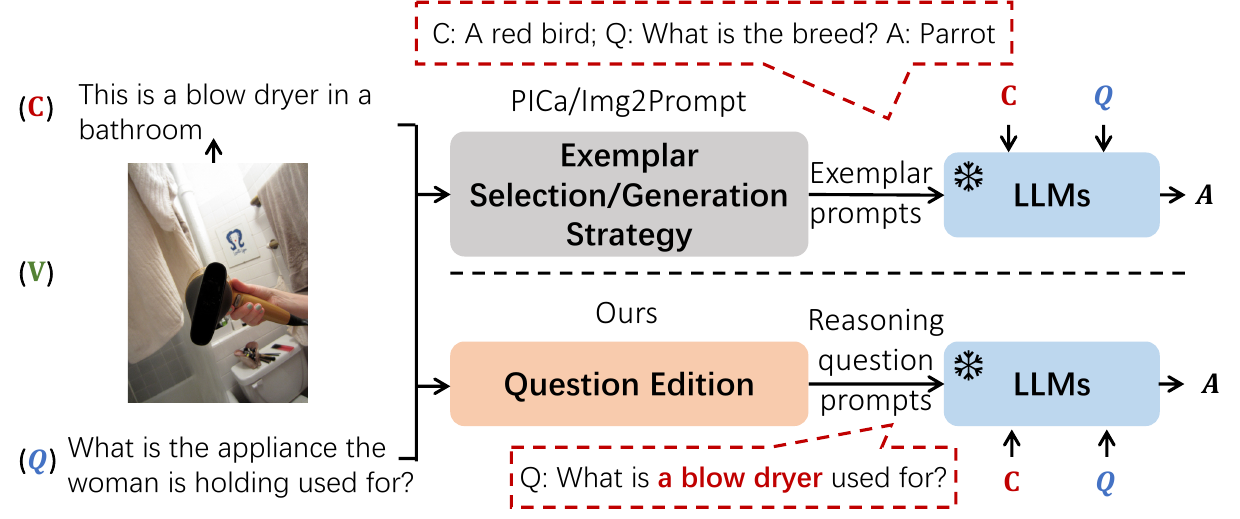}
    \caption{Comparison between existing prompting methods and our method on VQA tasks using frozen LLMs~\cite{brown:NeurIPS2020,zhang:arxiv2022}.
    The images are first converted into captions.
    Prior studies proposed different strategies to select exemplars from training data like \textsf{PICa}~\cite{yang:aaai2022} or generate synthetic exemplars like Img2Prompt~\cite{guo:cvpr2023}.
    In contrast, our method focuses on question prompt generation, where self-contained questions are produced in an unsupervised manner such that LLMs can easily capture the intent of questions and fully exert their potential.
    }
    \vspace{0.5cm}
    \label{fig:motivation}
\end{figure}

Visual Question Answering (VQA) tasks require a system to answer a textual question about an image. 
Diverse studies focused on solving visual questions, the answer of which can be directly derived from the image\cite{antol:2015iccv}, or questions requiring outside knowledge beyond the image content~\cite{wang:arXiv2022,marino:CVPR2019}.
Due to the enormous demand of manpower to annotate VQA datasets and the risk of human biases~\cite{agrawal:cvpr2018,changpinyo:arXiv2022}, there are quite a few studies proposing methods to solve zero-shot VQA tasks, where no image-question pair is provided for training~\cite{Song:ACL2022,banerjee:arXiv2020,jiang:arXiv2023}.

To solve zero-shot VQA tasks, early studies developed methods to synthesize training data so that conventional VQA models can be trained on the synthetic data~\cite{Song:ACL2022,jiang:arXiv2023,changpinyo:arXiv2022,chen:arXiv2023}.
Recently, Large Language Models (LLMs), which are trained on general text corpus, have shown excellent generalization capability on zero-shot tasks, such as information extraction~\cite{wei:arXiv2021} and logical reasoning~\cite{zhou:arXiv2022}.
Inspired by the intriguing properties of LLMs, \citeauthor{yang:aaai2022}~(\citeyear{yang:aaai2022}) first proposed \textsf{PICa}, which transfers images into captions then a frozen off-the-shelf LLM is applied to answer the question based on the caption context.
It not only saves the effort of pre-training a multi-modal model, but also provides world-knowledge to answer the questions.
Take the question in Figure~\ref{fig:motivation} as an example, the image is converted into the caption ``\textit{This is a blow dryer in a bathroom.}''.
The question ``\textit{What is the appliance the woman is holding used for?}'' should be answered with the caption as context.
To guide LLMs to better understand the tasks, in-context examples are selected from the training data as the prompts.
Soon after, another study proposes Img2Prompt~\cite{guo:cvpr2023}, which generates synthetic question-answer pairs via template-based and neural question-generation methods based on the images, which has shown impressive performance on zero-shot VQA benchmark datasets.

We observe that most of the existing prompting methods on VQA tasks focus on developing different exemplar selection/generation strategies to help LLMs better comprehend the task thus enhancing its capacity.
However, there is a demand of eliminating the semantic gap between captions and questions, which can be illustrated from two aspects:
(1) The current methods entirely rely on the understanding capability of LLMs to resolve the ambiguity and infer the intent of the questions, which might involve unexpected bias~\cite{kirk:nips2021,schramowski:nmi2022}.
As we can see, in Figure~\ref{fig:motivation}, the question asks about ``\textit{this appliance}'', which indicates ``\textit{blow dryer}''.
Due to the bias existing in LLMs, they may fail to parse the question correctly.
(2) 
LLMs are brittle to ill-posed questions, especially under the zero-shot setting.
In Figure~\ref{fig:motivation}, ``\textit{the woman is holding}'' is irrelevant to the image.
LLMs are sensitive to such noisy information and it may cause confusion to LLMs~\cite{zhao:ICML2021}.
In this case, disambiguating the question is of high demand.

Motivated by the observation, we present \textbf{RQ} prompts, which are \textbf{R}easoning \textbf{Q}uestion prompts for improving the understanding capability of LLMs under zero-shot VQA scenarios.
Specifically, we design an unsupervised question edition module to convert original questions into self-contained questions by editing the segments of the question.
We propose a search algorithm to generate the possible edited questions and rank them by a scoring function.
The scoring function measures sentence fluency, semantic integrity and syntactic invariance.
Eventually, the top-ranked reasoning question prompts are utilized to generate a set of candidate answers.
Following the heuristic prompting in Prophet~\cite{shao:arXiv2023}, where prompting is divided into answer generation and answer choosing steps, we encode both answer candidates and a confidence score to form answer heuristics for choosing.
The confidence score takes both the confidence of the reasoning question prompt and the generated answer into consideration, which produces a comprehensive score for choosing.
Our contribution can be summarized as follows:
\begin{itemize}
    \item We propose RQ prompts, which aim to improve zero-shot VQA tasks via LLMs by providing edited questions as prompts.
    No extra data as well as supervision is needed for the RQ prompts generation procedure.
    \item We design a novel confidence scoring function for the answer heuristics, which can comprehensively measure the answer candidates.
    \item 
    Reasoning question prompts generally improve existing baselines with absolute improvement ranging from $0.3$ to $5.2$ points.
    Our method achieves new state-of-the-art results on three out of four evaluated zero-shot VQA data sets.
\end{itemize}
\section{Related Work}
\label{sec:related}

\begin{figure*}[t] 
	\centering
	\includegraphics[scale = 0.6, trim={0.1cm 0.3cm 0.5cm 0.2cm}, clip]{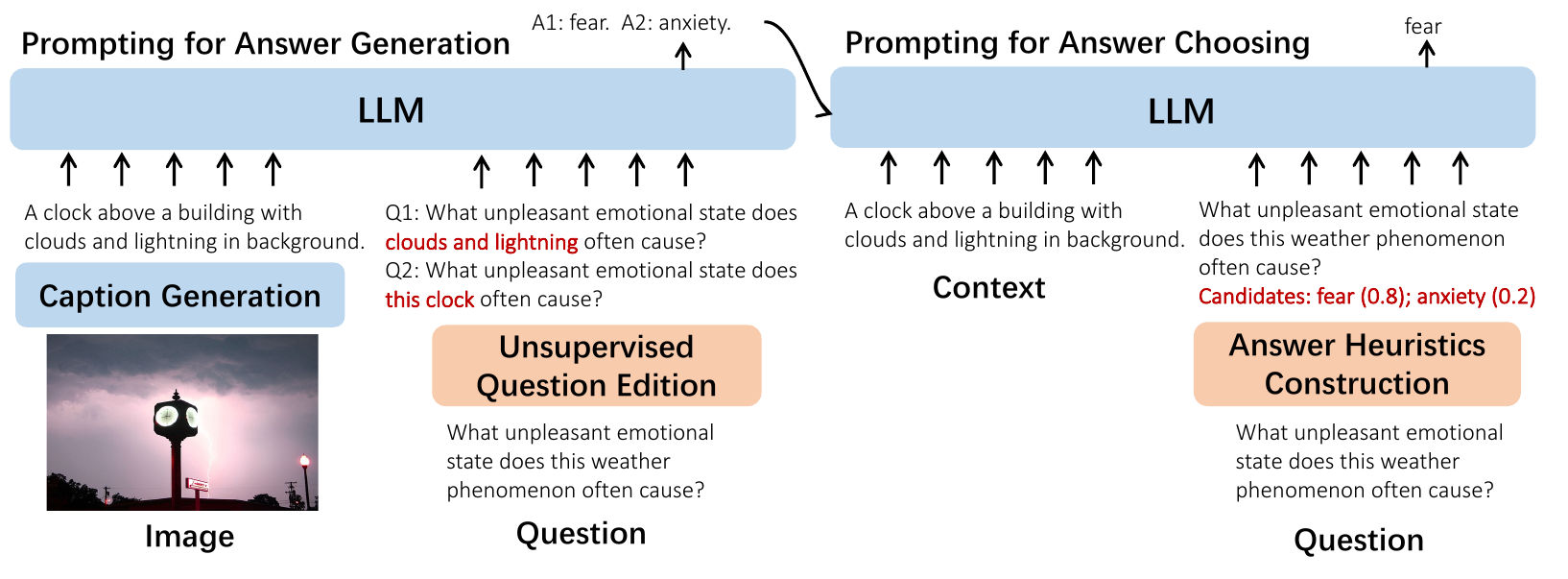} 
    \vspace{-0.3cm}
	\caption{The illustration of our prompting method that enables LLMs to perform VQA tasks with two-step reasoning. 
    The blue blocks denote the modules with frozen parameters and the orange blocks denote the modules we propose to generate reasoning question prompts and answer heuristics.}
	\label{fig:method}
\end{figure*}


\subsection{VQA tasks}
Given a textual question, VQA tasks require a system to answer the question by decoding the information from an image and even utilizing external knowledge.
Several benchmark datasets~\cite{wang:arXiv2015,wang:arXiv2022,marino:CVPR2019,schwenk:ECCV2022}, including complex reasoning questions, facilitate the development of this field.
To incorporate with external knowledge, early methods turned to textual Knowledge Bases (KBs) and applied either graph-based~\cite{Narasimhan:NeurIPS2018,Zhu:IJCAI2020,Ziaeefard:COLING2020,Li:AAAI2022} or transformer-based approaches~\cite{Gardres:emnlp2020,gao:CVPR2022} to introduce the KB information into the question answering module.
Besides, multi-modal KBs are also leveraged to solve VQA tasks.
\citeauthor{wu:AAAI2022}~(\citeyear{wu:AAAI2022}) combine Wikipedia, ConceptNet and Google images to supplement multi-modal knowledge.
With the emergence of language models, researchers consider them as implicit KBs~\cite{yang:aaai2022,shao:arXiv2023} and there are several studies~\cite{marino:CVPR2021,gui:arXiv2021,lin:arXiv2022,garcia:WWW2022} combining explicit and implicit knowledge to improve model's ability of handling visual questions.
Recently, large language models impress people by their quantum leap of understanding and reasoning capabilities.
Several studies~\cite{yang:aaai2022,shao:arXiv2023} reformulate VQA tasks into a textual question answering task by converting the images into captions and apply in-context learning to activate the implicit knowledge in LLMs~\cite{brown:NeurIPS2020}.
In this paper, we discuss VQA tasks under zero-shot scenarios, which brings in new challenges to the tasks.

\ignore{
\lxadd{Given a textual question, the task needs to answer it through selecting relevant information from an image and external knowledge. Several created benchmarks~\cite{wang:arXiv2015,wang:arXiv2022,marino:CVPR2019,schwenk:ECCV2022}, containing complex reasoning questions, facilitate the development of this field. Most of the early methods~\cite{narasimhan:ECCV2018,Ouyang:NeurIPS2022,Zhu:IJCAI2020,Ziaeefard:COLING2020,Li:AAAI2022,Gardres:emnlp2020,gao:CVPR2022} focused on textual knowledge base (KB), which involve graph-based approaches~\cite{Narasimhan:NeurIPS2018,Zhu:IJCAI2020,Ziaeefard:COLING2020,Li:AAAI2022} and transformer-based approaches~\cite{Gardres:emnlp2020,gao:CVPR2022}. Later, multi-modal KB arouses the interest of researchers~\cite{wu:AAAI2022,ding:CVPR2022}. For example, ~\cite{wu:AAAI2022} combines textual (Wikipedia, ConceptNet) and visual (Google images) knowledge to provide multi-modal information. With the emergence of large language models (LLMs) , implicit knowledge in LLMs is regarded as a new form of KB~\cite{yang:aaai2022,shao:arXiv2023}. Practically, ~\cite{yang:aaai2022} transforms the image into textual expressions and uses in-context learning to activate implicit knowledge in GPT-3~\cite{brown:NeurIPS2020}. This inspires ~\cite{marino:CVPR2021,gui:arXiv2021,lin:arXiv2022,garcia:WWW2022} to leverage both explicit and implicit KB to improve model's ability of handling challenging questions.  In this study, our proposed idea benefits not only VQA, but also knowledge-based VQA, which helps to demonstrate its generalization.
}
}

\subsection{Zero/Few shot of VQA tasks}

There is a line of work focusing on solving zero/few-shot VQA tasks.
A general solution is to augment image-question pairs for training.
Multi-modal pre-training models like CLIP~\cite{radford:icml2021} are frequently leveraged to generate synthetic question-answer pairs from images~\cite{banerjee:arXiv2020,changpinyo:arXiv2022}.
After that, a VQA model can be trained with the augmented data so that it can learn patterns and answer questions in the test set.
\citeauthor{tsimpoukelli:NeurIPS2021}~(\citeyear{tsimpoukelli:NeurIPS2021}) simply train a vision encoder to represent each image as a sequence of continuous embeddings, which could collaborate well with a frozen language model.
This inspires more studies ~\cite{manas:arXiv2022,alayrac:nips2022,liang:arXiv2022,jiang:arXiv2023} proposing parameter-efficient methods to combine both pre-trained vision models and language models for zero/few-shot VQA.
\citeauthor{guo:cvpr2023}~(\citeyear{guo:cvpr2023}) shift to the paradigm of leveraging LLMs to solve VQA tasks, they propose a method to automatically generate prompts as exemplars under the zero-shot setting.
This is the closest study to our work, but our work is different as we focus on the question prompts instead of exemplar prompts.

\subsection{Prompt Tuning of LLMs}

Prompts are significant regarding the inference of LLMs.
It helps guide the LLMs to activate the potentials of understanding and reasoning.
A question could be part of the prompt.
It should be well designed to fit the nature of the evaluated tasks~\cite{schick:arXiv2020}.
For example, pattern-verbalizer pair is one type of question prompt which maps diverse tasks into a word prediction task.
Besides, there are some other prompts.
An instructional prompt primarily contains a natural language description of the underlying task.
Generally, a narrative sentence is annotated manually as the instruction prompt~\cite{wang:emnlp2022,mishra:acl2022}.
Recently, researchers decompose a complex task into sub-tasks so that the multiple instruction prompts guide LLMs to handle sub-tasks step by step~\cite{wei:arXiv2022,wang:arXiv2022,Kojima:arXiv2022,zhang:arXiv2022a,zhou:arXiv2022}.
An exemplar prompt guides LLMs by showing some examples from the training data.
There are a number of studies proposing different strategies to select or generate good exemplar prompts for LLMs~\cite{brown:NeurIPS2020,min:arXiv2022,kim:arXiv2022,liu:arXiv2021}.
Instead of discrete text, prompts could be in the format of continuous embeddings, researchers have developed diverse methods to learn better embeddings~\cite{qin:naacl2021,li:acl2021}.
Our work takes effort on improving the question prompt by eliminating the semantic gap between the original question and images for zero-shot VQA tasks.

\ignore{
\lxadd{
Since \cite{brown:NeurIPS2020} designed the in-context learning (ICL) and ~\cite{schick:arXiv2020} formulated a mature framework (pattern-verbalizer pair, PVP) , prompt tuning has received increasing attention. Specifically, pattern is a certain function which changes the original sentence to a new one that contains a mask token, and verbalizer is another function which maps the LLMs' prediction to a word from the desired vocabulary. As LLMs shows its great few-shot and zero-shot capabilities, three key prompt tuning techniques for LLMs are frequently applied. Firstly, ICL ~\cite{brown:NeurIPS2020,min:arXiv2022,kim:arXiv2022,liu:arXiv2021,lu:arXiv2021,chen:arXiv2022,chen:arXiv2021,zhao:ICML2021} chooses a few training examples as the auxiliary context input. Although ICL is naturally suitable for few-shot scenarios, it exists indirect ways to generate pseudo examplars for zero-shot scenarios~\cite{guo:cvpr2023}. Secondly, instruction-tuning ~\cite{mishra:acl2022,wei:arXiv2021,thoppilan:arXiv2022l,Ouyang:NeurIPS2022} constructs a set of task instructions and teaches LLMs to response rightly according to current task instruction. Thirdly, chain of thought (CoT) ~\cite{wei:arXiv2022,wang:arXiv2022,Kojima:arXiv2022,zhang:arXiv2022a,zhou:arXiv2022} refers to a series of intermediate reasoning steps that guide LLMs to adopt the strategy of divide and conquer. However, previous work seldomly considers making better description of the original question. Therefore, our work plan to do it.
}
}
\section{Methods}
\label{sec:method}

\subsection{Overview}

In this section, we introduce our prompting method for solving zero-shot VQA tasks.
Following Prophet~\cite{shao:arXiv2023}, which is a heuristic prompting framework, we also decompose the task into two steps as shown in Figure~\ref{fig:method}.
In prompting step for answer generation, we convert an image into a caption with a frozen caption model~\cite{zhang:cvpr2021} as the context of the given question.
Particularly, we edit the question with an unsupervised method, namely \textbf{Unsupervised Question Edition} module, to transfer the original question into the reasoning question prompts.
For each reasoning question prompt, we generate a candidate answer from a frozen LLM.
In prompting step for answer choosing, we construct answer heuristics via \textbf{Answer Heuristics Construction} module based on the candidate answers generated above.
Then a frozen LLM is required to choose correct answer among these candidates.
Each candidate answer in the prompt is associated with a confidence score taking account of the confidence of both question prompts and answers.

\subsection{Prompting for Answer Generation}

To bridge the gap between the image captions and questions, we generate reasoning question prompts to avoid errors resulting from missing reasoning step.
Then we generate candidate answers based on them.
We define the reasoning question prompts should meet the following criterias:

\begin{itemize}
    \item The generated questions should not contain any ellipsis and ambiguity.
    In other words, they should be self-contained.
    Such that LLMs could easily understand the question without guessing the implicit information.
    Take the question in Figure~\ref{fig:method} as an example, ``\textit{this weather phenomenon}'' in question should be explicated by ``\textit{clouds and lightning}''.
    \item The self-contained question should be produced in the absence of supervision signal under a zero-shot setting.
    A neural network-based model is difficult to be applied as it requires a large volume of labeled data to learn how to generate a self-contained question. 
\end{itemize}

To meet the above criterias, we propose an unsupervised method to edit the original question with the consideration of its image caption.
There are two advantages of conducting edition on the original questions instead of generation:
(1) It is controllable to revise the original questions by substitution. 
Only segments of the questions can be changed and the major semantics of the original questions is maintained.
(2) Even without parallel labeled data, it is possible to conduct edition on the original question by a search algorithm holding a search objective.
On this basis, we design an unsupervised question edition module to convert the original question into a reasoning question prompt.

\subsubsection*{\textbf{Unsupervised Question Edition}} 

\begin{figure*}[t] 
	\centering
	\includegraphics[scale = 0.47, trim={0.1cm 0.3cm 0.5cm 0.3cm}, clip]{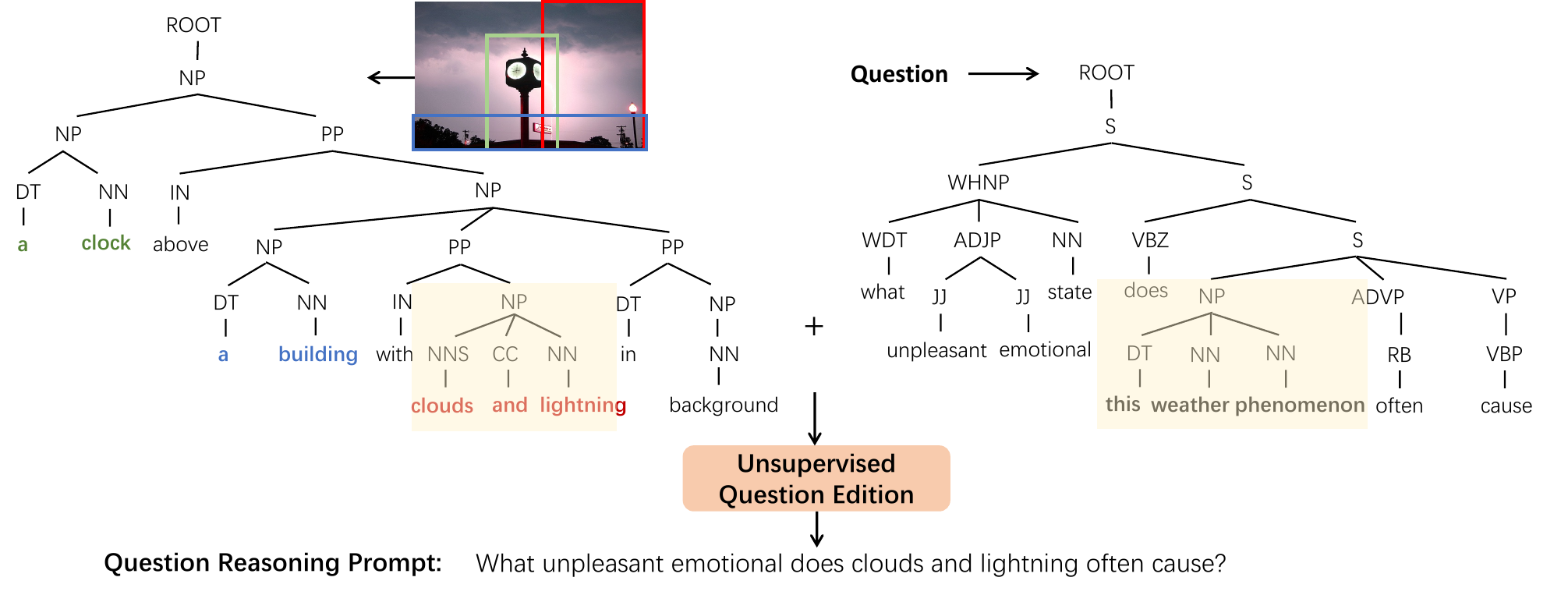} 
  \vspace{-0.3cm}
	\caption{The generation process of reasoning question prompts in unsupervised question edition module.
    Both the question and caption are transformed into constituency parse trees.
    The phrase-level constituents in the caption correspond to the different objects in the image, which are shown with different colors.
    They would be utilized to substituent segments of the original question to form a complete self-contained question.
    The yellow shades indicate that we substituent these constituents to form a reasoning question prompt.
    }
	\label{fig:parse}
\end{figure*}

Inspired from existing work on text simplification~\cite{kumar:acl2020}, we design an edit-based search algorithm to produce the reasoning question prompts by conducting substitution operations on the constituency parse tree.
As shown in Figure~\ref{fig:parse}, ``\textit{clouds and lightning}'' (NP) and ``\textit{this weather phenomeon}'' (NP) are both considered as phrase-level constituents with the same root tag ``\textit{Noun Phrase}'' based on the constituency parse trees.
By replacing ``\textit{this weather phenomenon}'' with ``\textit{clouds and lightning}'', we can obtain a self-contained question ``\textit{what unpleasant emotional does clouds and lightning often cause?}''.

Given a caption and a question, our search algorithm iteratively performs edits to search for a candidate.
Specifically, starting from the constituents of the caption, we consider all the constituents of the original question and conduct substitution to generate candidates.
Each candidate will be measured by a scoring function considering the sentence fluency, semantics integrity and syntactic invariance.
The candidate with the score higher than a threshold can be saved and further edited.
The detailed search algorithm is displayed in Algorithm~\ref{alg:search}.


\begin{algorithm}
\caption{Search Algorithm of Reasoning Question Prompts}
\begin{algorithmic}[1]
\Procedure{QE}{$C$, $Q$} \Comment{$C$ and $Q$ are the parse trees of captions and original questions, respectively.}
\State $\mathcal{S} = \{Q\}, \mathcal{S}_{batch} = \{Q\}$
\For {$j = 1, ..., l$} \Comment{$l$: number of constituents in $C$}
    \State $\mathcal{S}_{best} = \{\}$ \Comment{Initialization}
\For {$Q'$ in $\mathcal{S}_{batch}$}
  \For {$i = 1, ..., n$} \Comment{$n$: number of constituents in $Q$}
    \State $\tilde{Q} \gets substituent(Q'[i], C[j])$ \Comment{Edit constituents}
    \State $s \gets f(\tilde{Q})$ \Comment{Score the above candidate}
    \If {$s > (f(Q) - \rho)$}
    \State $\mathcal{S}_{best} \gets \mathcal{S}_{best} \cup \{\tilde{Q}\}$ \Comment{Save the candidate}
    \EndIf
    \EndFor
  \EndFor
\State $\mathcal{S}_{batch} \gets \mathcal{S}_{best}$
\State $\mathcal{S} \gets \mathcal{S} \cup \mathcal{S}_{best}$
\EndFor
\State \textbf{return} $\mathcal{S}$
\EndProcedure
\end{algorithmic}
\label{alg:search}
\end{algorithm}

Next, we present our scoring function. 
To evaluate the quality of the candidate, we consider the following aspects comprehensively:

\begin{itemize}
    \item \textbf{LM Score.} We employ a probabilistic language model (LM) to measure the language fluency of a candidate, which is widely applied in unsupervised text compression and simplification tasks~\cite{kann:CoNLL2018,miao:aaai2019}.
    As the training objective of LMs is to maximize the likelihood of sentences, a fluent sentence would have a higher joint probability, which can be denoted as $f_{LM}(\tilde{Q}) = \ln P_{LM}(\tilde{Q}) = \ln \prod_{i=1}^{T} P(w_i| w_{i-1}, ..., w_1)$, where $w_i$ is $i$-th token in $\tilde{Q}$ and $T$ is the length of the sentence.
    \item \textbf{Semantic Integrity.} To avoid the dramatic change to the semantics of the original question after edition, we employ cosine similarity to measure the meaning preservation, where the sentence embedding is computed as the weighted average of tokens in sentences.
    We denote it as $f_{Semantic}(\tilde{Q}) = cos(\tilde{Q}, Q)$.
    \item \textbf{Syntactic Invariance.} Since we would like to ensure the alternative constituents can hold the same syntactic attributes as the original one.
    This could maintain the syntactic structure of the original question and effectively avoid grammatical confusion.
    We identify whether the root tags of these constituents are same or not, which can be denoted as $f_{Syntactic}(\tilde{Q}) = \mathbb{I}(Tag_{Q[i]} = Tag_{\tilde{Q}[j]})$.
    Here $\mathbb{I}(\cdot)$ is an indicator function.
\end{itemize}

The overall scoring function is the product of the above aspects:
\begin{equation}
    f(\tilde{Q}) = f_{LM}(\tilde{Q})^{\alpha} f_{Semantic}(\tilde{Q})^{\beta} f_{Syntactic}(\tilde{Q}),
    \label{eq:fq}
\end{equation}
where the weights $\alpha$ and $\beta$ denote the importance of LM score and semantic integrity, respectively.
It is worth that syntactic invariance is a hard indicator function.
It only accepts the case when the replaced root tag is unchangeable so there is no importance weight needed.
As we can see, $f(\tilde{Q})$ is a scalar that indicates how likely $\tilde{Q}$ can act as a good reasoning question prompt for $Q$.
Eventually, we obtain a set $\mathcal{S}$ that contains $k$ reasoning question prompts.

\subsubsection*{\textbf{Prompt Design}}

With the generated $k$ reasoning question prompts, we construct the prompts for answer generation by concatenating the caption and each reasoning question prompts.
Following prior studies on prompt tuning~\cite{yang:aaai2022,guo:cvpr2023,shao:arXiv2023}, we construct the prompt with the consideration of instruction, context and questions:

\begin{itemize}[label={}, labelsep=0pt, leftmargin=10pt]
    \item \textbf{Instruction:} \texttt{Please answer the question according to the contexts}.
    \item \textbf{Context:} \texttt{[caption]}.
    \item \textbf{Question:} \texttt{[reasoning question prompt]}.
    \item \textbf{Answer:}
\end{itemize}

We will feed the $k$ prompts into LLMs in turn and greedy decoding on LLMs is performed on each prompt.
This results in $k$ candidate answers with their confidence scores.

In Figure~\ref{fig:method}, different reasoning question prompts capture different objects in the image such as ``\textit{clouds and lightning}'' and ``\textit{this clock}'', they can cover possible intents of the original question, which helps LLMs to decode answers with diverse reasoning paths.
This strategy has similar principle as Chain-of-Thought~\cite{Kojima:arXiv2022,zhang:arXiv2022a,zhou:arXiv2022}, which explicates the intermediate reasoning chains of the questions and makes it easier for LLMs to parse the question and do complicated reasoning.
After prompting for answer generation, we obtain two candidate answers, that are ``\textit{fear}'' and ``\textit{anxiety}'', which correspond to the two reasoning question prompts.

\subsection{Prompting for Answer Choosing}

Once we obtain multiple candidate answers, we construct prompts to let LLMs choose final answers among these candidate answers, which are known as heuristics-enhanced prompts in Prophet. 
This facilitates the LLMs to narrow down the range of answers.
We follow this strategy but define different confidence scores in Answer Heuristics Construction module.


\subsubsection*{\textbf{Answer Heuristics Construction}}

Starting from the generated candidate answers based on different reasoning question prompts, we define the confidence score of the candidate answer below:
\begin{equation}
    P(A) = \sum_{LLM(\tilde{Q}) \rightarrow A}P_(\tilde{Q})P_{LLM}(A|\tilde{Q}),
    \label{eq:pa}
\end{equation}
where $P{(\tilde{Q})}$ is the probability that we generate the $\tilde{Q}$ based on normalized $f{(\tilde{Q})}$ over $k$ prompts and $P_{LLM}(A|\tilde{Q})$ is the probability of the generated $A$ based on $\tilde{Q}$ via LLMs.
Since different reasoning question prompts may lead to the same answer, we can have $m$ candidate answers, where $m \le k$.
As we can see, the confidence score takes both confidences of question edition and answer generation into account, which comprehensively depicts the likelihood of a candidate answer for answering choosing.

\subsubsection*{\textbf{Prompt Design.}}
With the generated $m$ candidate answers, we construct the prompts for answer choosing by concatenating the caption, original question and candidate answers:

\begin{itemize}[label={}, labelsep=0pt, leftmargin=10pt]
    \item \textbf{Instruction}: \texttt{Please answer the question according to the contexts and candidates.}
    \item \textbf{Context}: \texttt{[caption]}.
    \item \textbf{Question}: \texttt{[original question]}.
    \item \textbf{Candidates}: \texttt{[$A_1$ P($A_1$)];[$A_2$ P($A_2$)];...;[$A_m$ P($A_m$)]}
    \item \textbf{Answer}:
\end{itemize}
where $P(A_m)$ denotes the confidence score for answer $A_m$, which reminds LLMs to focus more on the candidate answers with higher scores.
We consider the answer generated by this prompt as the final answer.

Compared with the two-stage prompting method of Prophet, our method is different in the way of generating and scoring answer candidates, which is rooted in our different motivation.
Prophet generates answer candidates by including frequent answers from training set, which is to replay the answer prediction in the training data.
Our method generates answer candidates by full-filling the original questions with possible intents, which is to shorten the semantic gap between images and questions under the zero-shot setting.
It is worth noting that even though the prompting method is designed for the zero-shot VQA task, we can still insert in-context examples behind the instructional prompt if it is needed.

\section{Experiments}
\label{sec:exp}

\begin{table*}
  \caption{Zero-shot evaluation on VQAv2, OK-VQA, and A-OKVQA.
  The first section contains zero-shot methods with LLMs which utilize no training data but may synthesize some exemplars.
  The middle section contains zero-shot methods with end-to-end training on other multi-modal data.
  The last section contains few-shot methods with LLMs.
  The numbers in brackets denote the improvement gain brought by our reasoning question prompts. 
  The results with $\diamond$ denote the baselines we implement methods by ourselves.
  Otherwise, we copy results from their original papers.
  } 
  \label{tab:main}
  \begin{tabular}{l | c c c | c c c c}
    \toprule
    Method & Model & Shot & Examplar & OK-VQA & VQAv2 & \multicolumn{2}{c}{A-OKVQA}  \\
    & size & number & number & test & val & val & test \\
    \midrule
    \multicolumn{8}{c}{\textit{Zero-shot Evaluation with Frozen LLMs}} \\
    \textsf{PICa}$_{\text{ \{GPT-3\}}}$ & $175$B & $0$ & $0$& $17.7$ & $-$ & $23.8^\diamond$ & $-$\\
    Img2Prompt$_{\text{ \{OPT\}}}$ & $6.7$B  & $0$ & $30$ & $38.2$ & $52.2^\diamond$ & $33.3$ & $32.2$ \\
    Img2Prompt$_{\text{ \{OPT\}}}$ & $30$B  & $0$ & $30$ & $41.8$ & $54.2^\diamond$ & $36.9$ & $33.0$ \\
    Img2Prompt$_{\text{ \{GPT-3\}}}$ & $175$B  & $0$ & $30$ & $42.8$ & $-$ & $38.9^\diamond$ & $43.4^\diamond$\\
    Img2Prompt$_{ \text{ \{OPT\}}}$ & $175$B  & $0$ & $30$ & $45.6$ & $\mathbf{60.6}$ & $42.9$ & $40.7$ \\
    \textsf{PICa}+RQ prompt$_{\text{ \{GPT-3\}}}$ (Ours) & $175$B  & $0$ & $0$ & $20.3(\uparrow 2.6)$ & $-$ & $29.0(\uparrow 5.2)$ & $-$\\
    Img2Prompt+RQ prompt$_{\text{ \{OPT\}}}$ (Ours) & $6.7$B  & $0$ & $30$ & $38.5(\uparrow 0.3)$& $52.9(\uparrow 0.7)$ & $36.3(\uparrow 3.0)$ & $31.5$\\
    Img2Prompt+RQ prompt$_{\text{ \{OPT\}}}$ (Ours) & $30$B  & $0$ & $30$ & $42.1(\uparrow 0.3)$ & $54.5(\uparrow 0.3)$ &  $38.1(\uparrow 1.2)$ & $35.2(\uparrow 3.0)$\\
    Img2Prompt+RQ prompt$_{\text{ \{GPT-3\}}}$ (Ours) & $175$B  & $0$ & $30$ & $\mathbf{46.4}(\uparrow 3.6)$ & $-$ & $\mathbf{43.2}(\uparrow 4.3)$ & $\mathbf{43.9}(\uparrow 0.5)$\\
    \midrule
    \multicolumn{8}{c}{\textit{Zero-shot Evaluation with Pre-trained VQA methods}} \\ 
    VL-T5$_{ \text{ \{no-vqa\}}}$ & $224$M  & $0$ & $0$ & $5.8$ & $13.5$ & $-$ & $-$ \\
    FewVLM$_{ \text{ \{large\}}}$ & $740$M  & $0$ & $0$ & $16.5$ & $47.7$ & $-$ & $-$ \\
    VLKD$_{ \text{ \{ViT-L/14\}}}$ & $408$M  & $0$ &  $0$ & $13.3$ & $44.5$ & $-$ & $-$\\
    Frozen & $7$B  & $0$ & $0$ & $5.9$ & $29.5$ & $-$ & $-$\\
    Flamingo & $80$B  & $0$ & $0$ & $50.6$ & $-$ & $-$ & $-$\\
    \midrule 
    \multicolumn{8}{c}{\textit{Few-shot Evaluation with Frozen LLMs}} \\  
    \textsf{PICa}$_{\text{ \{GPT-3\}}}$ & $175$B  & $16$ & $16$ & $46.5$ & $54.3$ & $-$ & $-$ \\
    Prophet$_{\text{ \{GPT-3\}}}$ & $175$B  & $20$ & $20$ & $61.1$ & $-$ & $-$ & $-$ \\
  \bottomrule
\end{tabular}
\end{table*}

In this section, we evaluate reasoning question prompts on zero-shot VQA tasks and compare with existing methods.
Furthermore, we perform comprehensive analysis to interpret its performance under different scenarios.
We also conduce ablation study on important design choices and show some qualitative examples. 

\subsection{Experimental Setup}

\subsubsection*{\textbf{Datasets}}

We evaluate reasoning question prompts on \textbf{OK-VQA}~\cite{marino:CVPR2019}, \textbf{A-OKVQA}~\cite{schwenk:ECCV2022} and \textbf{VQAv2}~\cite{goyal:cvpr2017}, which contains image-question pairs that are derived from COCO datasets\cite{lin:eccv2014}.
The questions in these datasets require perception to the image.
Some of them even require commonsense beyond the image to answer.
Specifically, OK-VQA\footnote{\url{https://okvqa.allenai.org/}} contains $5,046$ test questions.
A-OKVQA\footnote{\url{https://allenai.org/project/a-okvqa/home}} contains $1,100$ and $6,700$ questions for validation and testing, respectively.
VQAv2\footnote{\url{https://visualqa.org/download.html}} is a large dataset,
We leverage the validation set of VQAv2 for evaluation, which contains $214,354$ questions.
For evaluation measurement, we follow their official evaluation metrics to measure the performance.

\subsubsection*{\textbf{Comparable Methods}}

As our reasoning question prompts can collaborate with any LLMs, we evaluate our methods with different LLMs as backbones.
Notably, existing methods like \textsf{PICa} and Img2Prompt are prompting methods to provide exemplars prompts for VQA tasks. 
We consider their methods as baselines then include our reasoning question prompts and observe if there is any performance improvement brought by the involvement of our method.

Besides, we compare our method with other pre-trained zero-shot VQA methods, such as Flamingo~\cite{alayrac:nips2022}, Frozen~\cite{tsimpoukelli:nips2021} VL-T5~\cite{cho:icml2021}, FewVLM~\cite{jin:acl2022} and VLKD~\cite{dai:acl2022}.
These methods aim to propose different pre-trained multi-modal models on large-scale vision-language datasets, which can be easily adapted to new VQA challenges without training.

\subsubsection*{\textbf{Implementation Details}}

For the LM used in the unsupervised question edition module, we leverage a pre-trained LM model from existing work, which is a two-layer, 256 dimensional recurrent neural network with gated recurrent unit (GRU)~\cite{kumar:acl2020} fine-tuned on OK-VQA test set.
Compared with LMs like BERT and Roberta, this model is enriched with syntactic information, which is more suitable for our method.
$\alpha$ and $\beta$ in Equation~(\ref{eq:fq}) are set to $0.3$ and $1$, respectively.
We set $\rho$ as $0.5$ to avoid overwhelming question reasoning prompts.
If there is a maximum limit number for the generated reasoning question prompts, we sort all candidate reasoning question prompts and select the top-$k$ based on their scores.
More details about the unsupervised edition module can be found in Appendix~\ref{ap:edition}.

Regarding LLMs, to show the generalization capability of our reasoning question prompts, we conduct experiments on different LLMs with different sizes, including open source OPT\footnote{\url{https://huggingface.co/docs/transformers/model_doc/opt}}, GPT-3\footnote{\url{https://openai.com}} and BLOOM.\footnote{\url{https://huggingface.co/docs/transformers/model_doc/bloom}}.
Regarding different baselines, such as Img2Prompt\footnote{\url{https://github.com/salesforce/LAVIS/tree/main/projects/img2llm-vqa}}, \textsf{PICa}\footnote{\url{https://github.com/microsoft/PICa}}, we follow their official implementation to convert images into captions via either VinVL-base pre-trained checkpoint\footnote{\url{https://github.com/pzzhang/VinVL}} or BLIP\footnote{\url{https://github.com/salesforce/BLIP}} and generate exemplar prompts via either CLIP\footnote{\url{https://github.com/OpenAI/CLIP}} or fine-tuned T5-large model\footnote{\url{https://github.com/google-research/text-to-text-transfer-transformer}}. 
Notably, we implement a light version of Img2Prompt on VQAv2 dataset due to our computation limitation, the details of which can be found in Appendix~\ref{ap:img2prompt}.

\subsection{Main Results}

We display our main results in Table~\ref{tab:main}. 
We have the following observations based on it:

\subsubsection*{\textbf{Overall effect of reasoning question prompts.}} Our reasoning question prompts can improve the performance of zero-shot VQA methods on most baselines.
The absolute improvement ranges from $0.3$ to $5.2$ points.
The largest gain is on A-OKVQA validation set with \textsf{PICa} baseline, where the absolute improvement is $5.2$ points.
This is a setting without any exemplar, which indicates the potential of reasoning question prompts under the scenarios with no access to any VQA data.
Even with some exemplars which are synthetically generated, reasoning question prompts can still improve the zero-shot VQA methods via LLMs.
As we can see, even with some syntactic exemplars generated by Img2Prompt model, there is general improvement from reasoning question prompts.
We observe the similar effect of RQ prompts with different LLMs, the results of which are displayed in Appendix~\ref{ap:diff_llm}.

\begin{figure*}
\centering
\begin{minipage}{.33\textwidth}
  \centering
  \includegraphics[width=\linewidth]{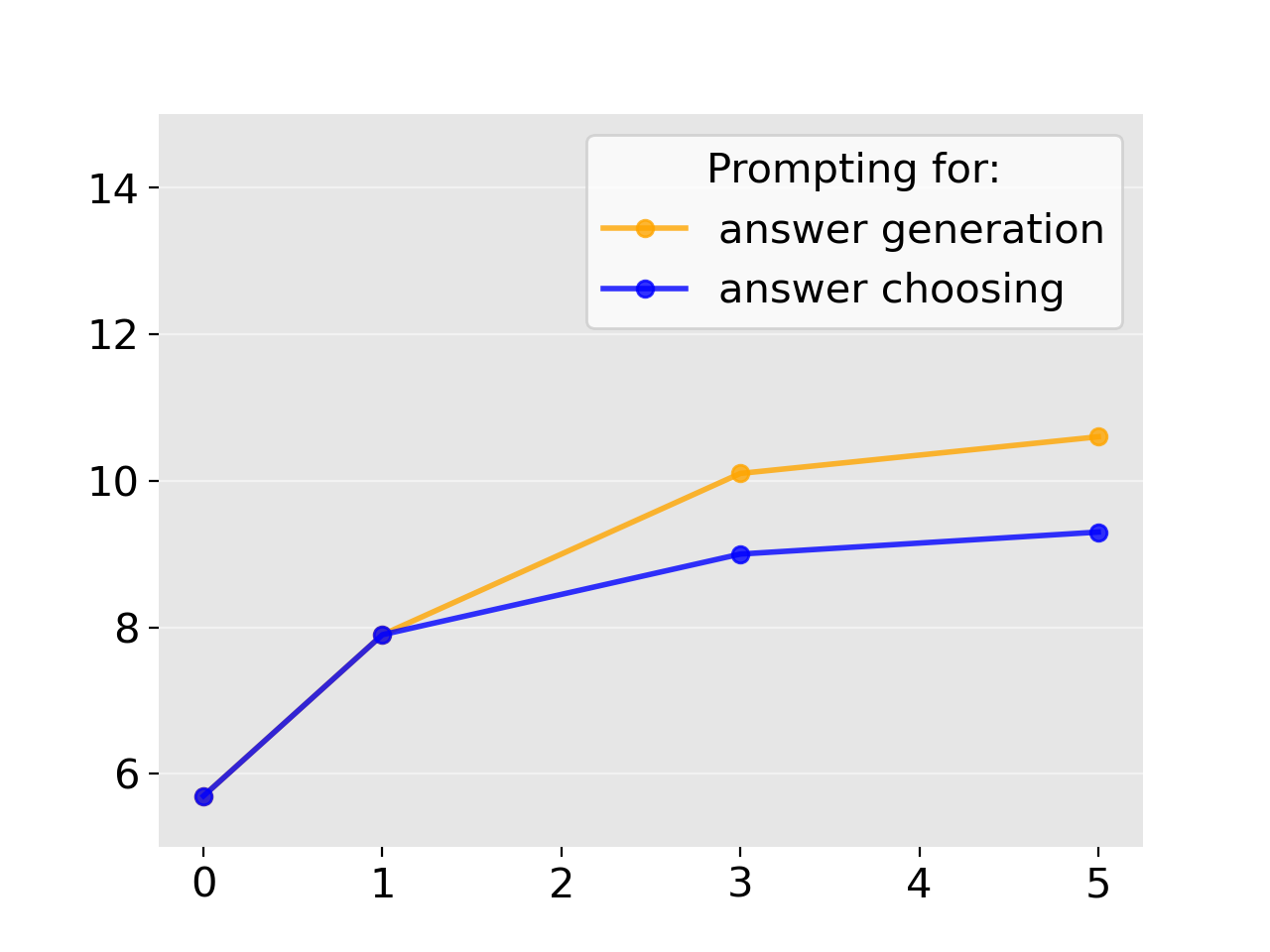}
  \vspace{-0.7cm}
  \subcaption{Shot number=0}
  \label{fig:test1}
\end{minipage}%
\begin{minipage}{.33\textwidth}
  \centering
  \includegraphics[width=\linewidth]{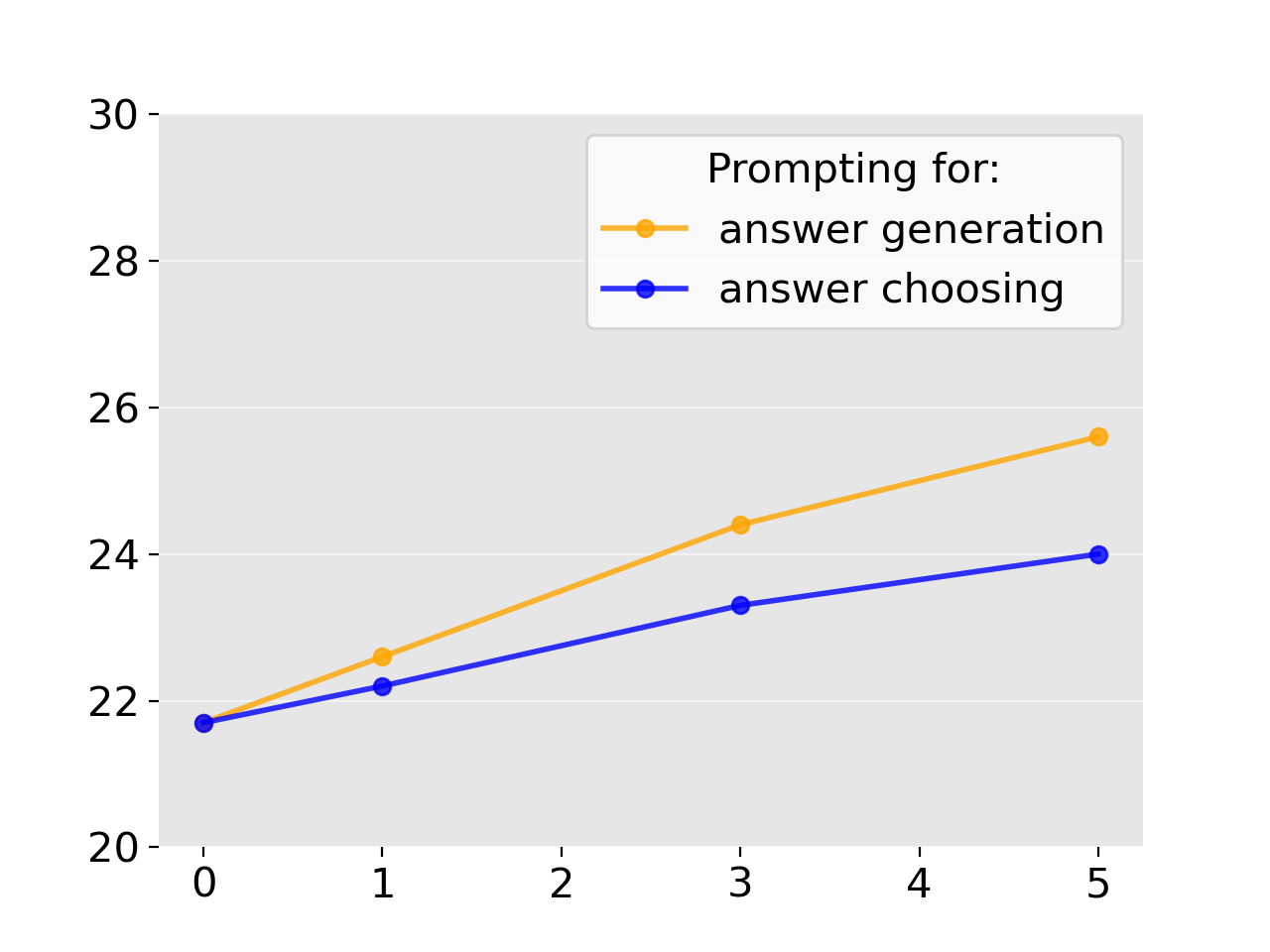}
  \vspace{-0.7cm}
  \subcaption{Shot number=1}
  \label{fig:test2}
\end{minipage}
\begin{minipage}{.33\textwidth}
  \centering
  \includegraphics[width=\linewidth]{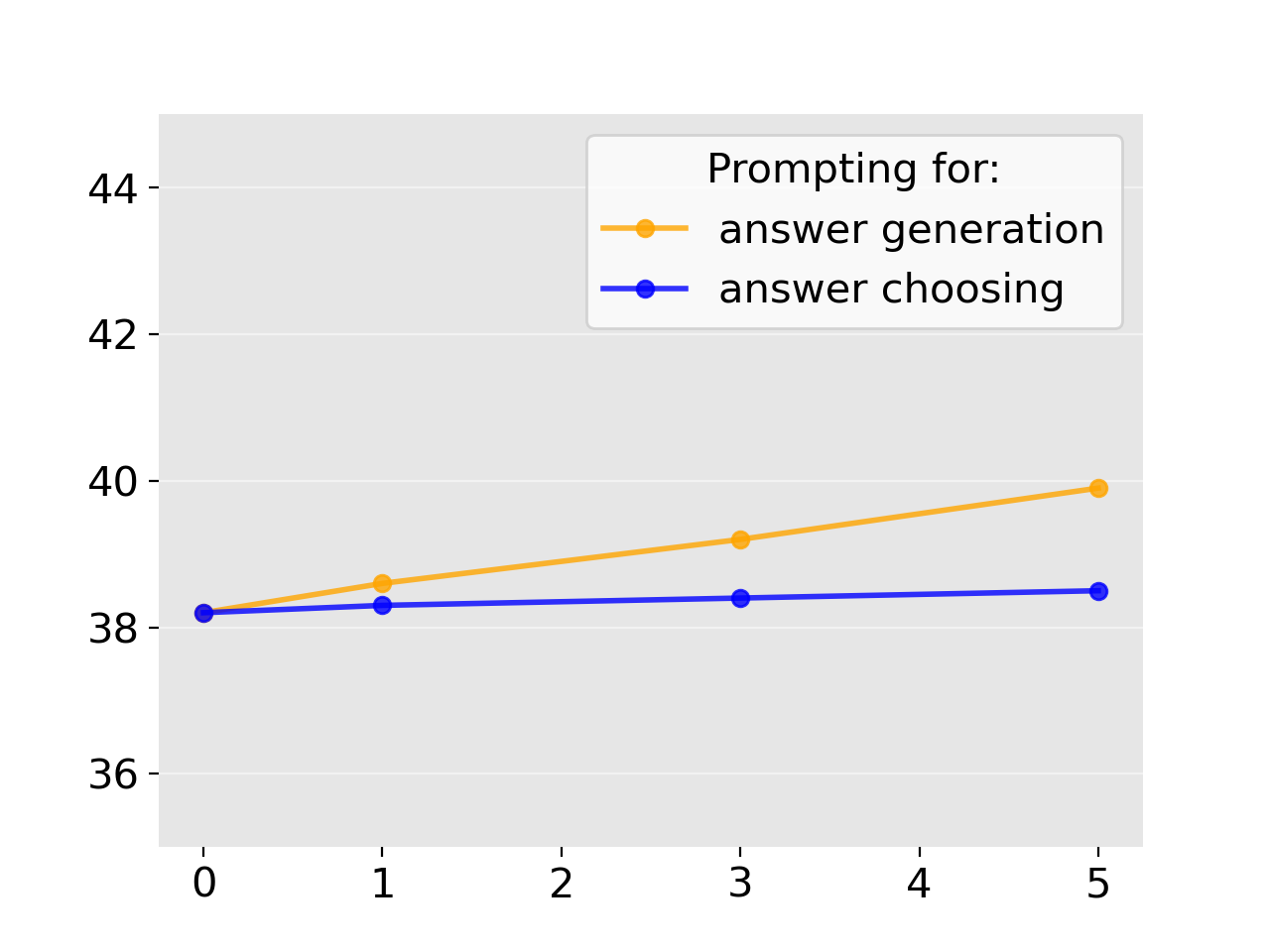}
  \vspace{-0.7cm}
  \subcaption{Shot number=16}
  \label{fig:test3}
\end{minipage}
\begin{minipage}{.33\textwidth}
  \centering
  \includegraphics[width=\linewidth]{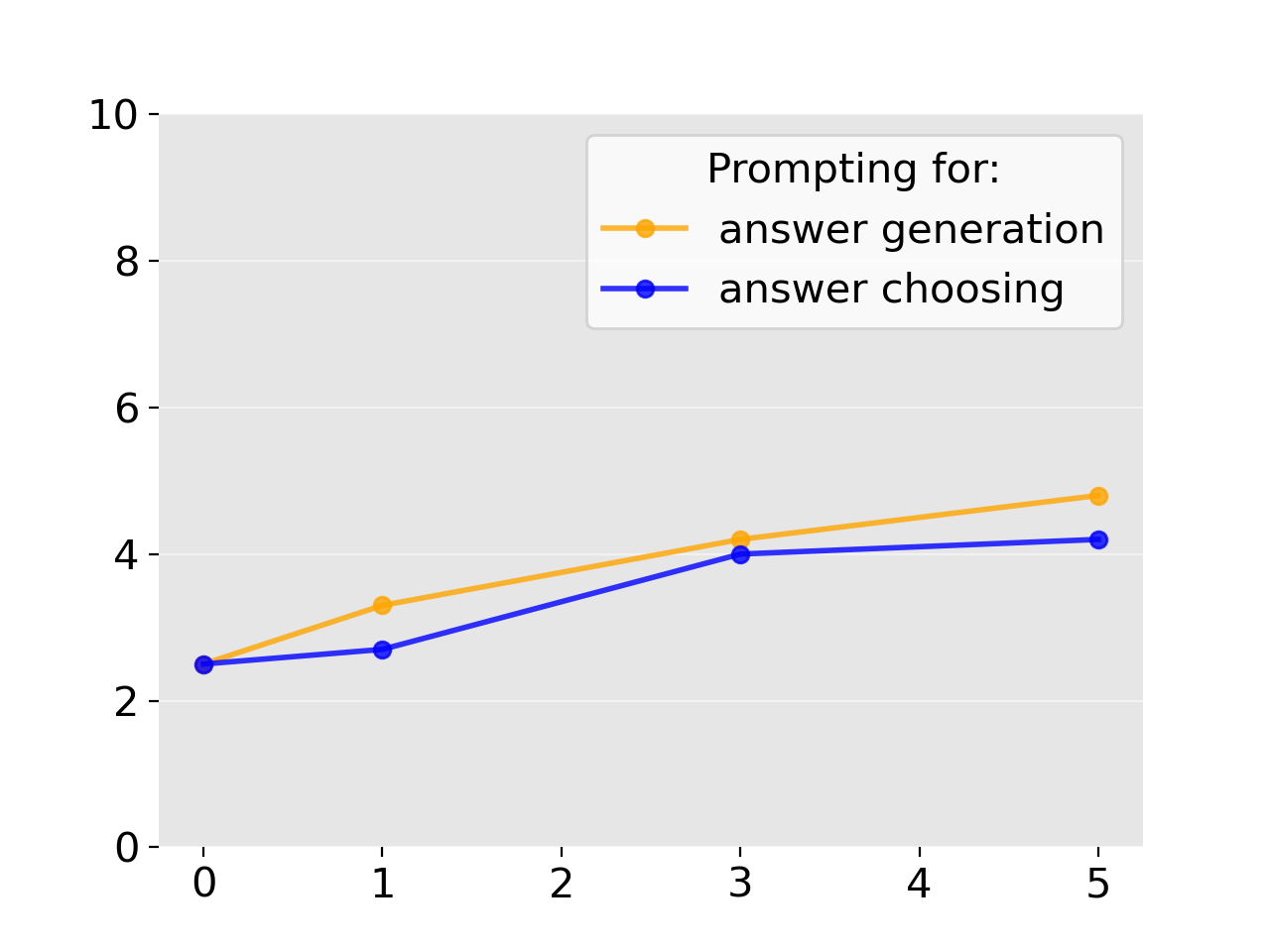}
  \vspace{-0.5cm}
  \subcaption{LLM=GPT-Neo 2.7B}
  \label{fig:test4}
\end{minipage}%
\begin{minipage}{.33\textwidth}
  \centering
  \includegraphics[width=\linewidth]{Figures/shot0.png}
  \vspace{-0.5cm}
  \subcaption{LLM=OPT 6.7B}
  \label{fig:test5}
\end{minipage}
\begin{minipage}{.33\textwidth}
  \centering
  \includegraphics[width=\linewidth]{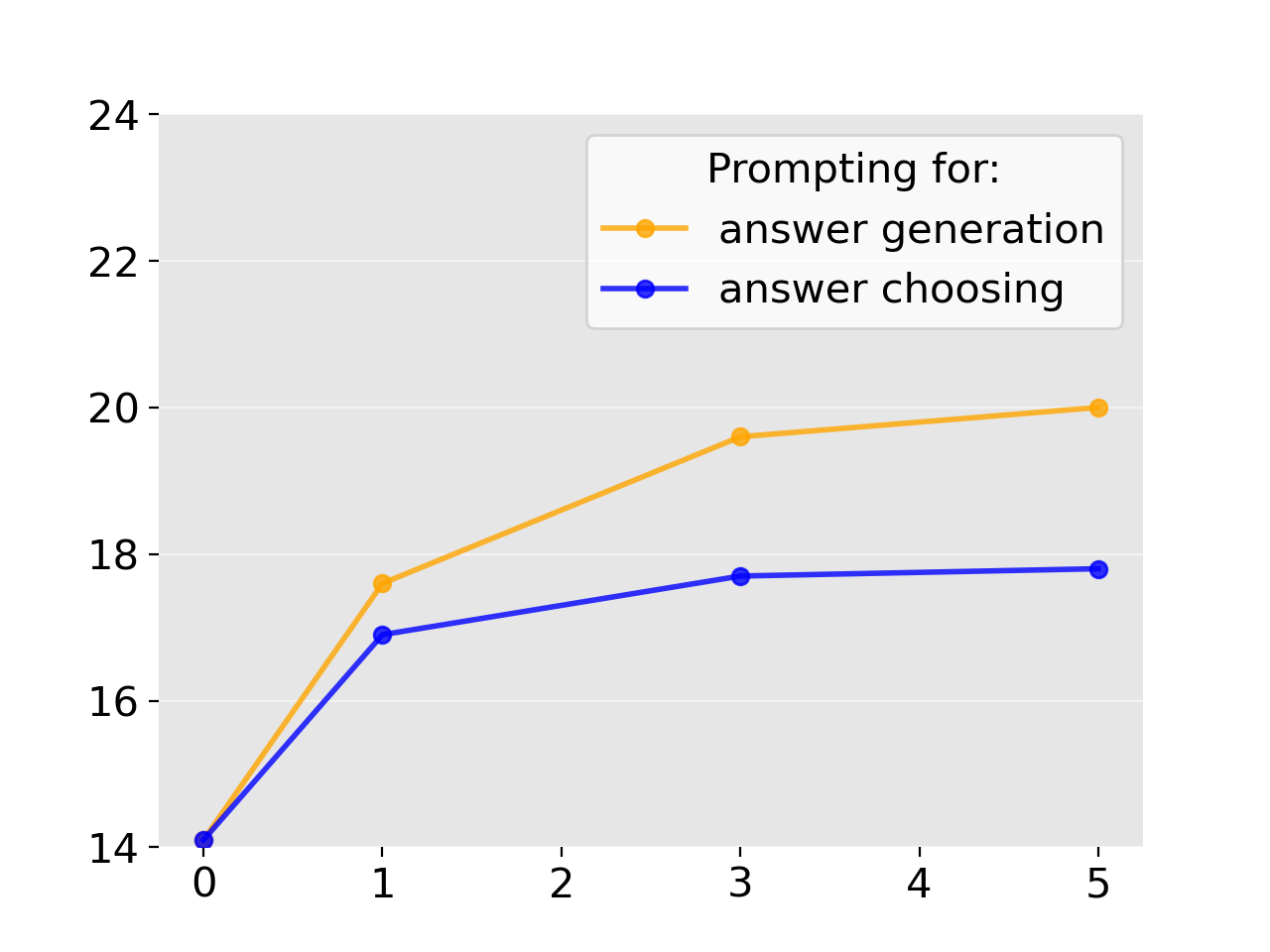}
  \vspace{-0.5cm}
  \subcaption{LLM=OPT 30B}
  \label{fig:test6}
\end{minipage}
\vspace{-0.3cm}
\caption{(a)-(c) denote evaluation of RQ prompts on the test set of OK-VQA having OPT 6.7B as the LLMs but with different shot numbers. 
(d)-(f) denote evaluation of RQ prompts on the test set of OK-VQA having shot numbers equal to $0$ but with different LLMs. 
We display results of prompting for answer generation and answer choosing.
X-axis denotes the value of $k$ and y-axis denotes the accuracy. 
When it comes to prompting for answer generation, we report the maximum accuracy among all the reasoning question prompts.}
\end{figure*}

\subsubsection*{\textbf{Comparison with other methods.}} 
Compared with existing zero-shot methods, reasoning question prompts with Img2Prompt$_{\{ \text{GPT-175B}\}}$ baseline can outperform all the existing zero-shot VQA methods and achieve the new state-of-the-art results on zero-shot evaluation with frozen LLMs on three out of four data sets.
Even though we cannot defeat Img2Prompt$_{\{ \text{OPT-175B}\}}$ on VQAv2 validation set, our reasoning question prompts can still bring in perfromance gain on our light re-implmentation results.
We notice that there are some competitive comparable methods like pre-trained VQA method Flamingo$_{\{ \text{80B}\}}$ and few-shot LLMs-based method Prophet$_{\{ \text{GPT-175B}\}}$, which lead to higher results than ours.
The former method is computationally expensive, which is pre-trained on billion-scale multi-modal datasets.
The latter one makes use of VQA training samples, which can obtain more guidance directly from the training data.

\subsection{More Analysis of RQ Prompts}

\subsubsection*{\textbf{Effect of $k$ for RQ Prompts on Different Shot Numbers.}}

\begin{figure*}[t] 
	\centering
	\includegraphics[scale = 0.5]{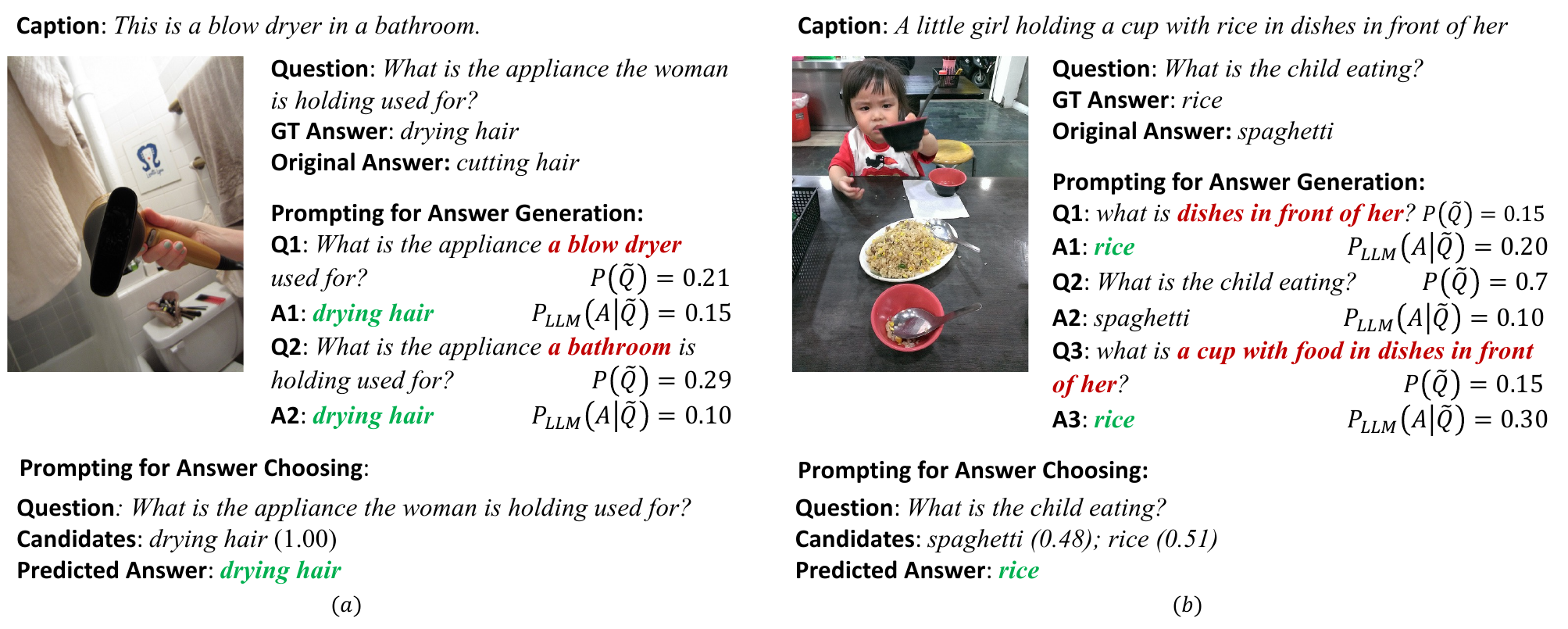} 
	\caption{Examples in A-OKVQA validation set the prediction of which are incorrect originally but correct with RQ prompts.}
	\label{fig:case}
\end{figure*}

As the results in Table~\ref{tab:main} have the mixed effect of shot number and LLMs, to analysis the effect of reasoning question prompts on different shot numbers, we control the other settings unchangeable and see how the performance changes with the increasing number of shot.
The results are displayed in Table~\ref{tab:step} (a)-(c).
As we can see, the performance gradually improves with the increasing $k$.
During answer generation, the more reasoning question prompts generated, the more likely that we can recall the correct answers.
We can observe the largest performance gain on the zero-shot setting.
This indicates that a reasoning question prompt is more likely to help when the guidance to the question is little.
Providing LLMs with self-contained questions, which explicate the intermediate reasoning to LLMs, can fully activate the potentials of LLMs.
When the shot number becomes $16$, the gain from reasoning question prompts becomes least visible.

\subsubsection*{\textbf{Effect of $k$ for RQ Prompts on Different LLMs.}}

We further control the shot number as $0$ and test the effect of $k$ for reasoning question prompts on LLMs with different parameter sizes.
The results are displayed in Table~\ref{tab:step} (d)-(f).
Similarly, the performance gain increases with the increasing $k$.
Furthermore, with the increase of model size, the performance gain becomes large.
Regarding GPT-Neo-2.7B, the performance increase brought by reasoning question prompts is not so obvious, which only has around $1$ point improvement.
Regarding OPT 30B, the performance increase brought by reasoning question prompts becomes around $3$ points.
This is because larger LLMs usually contain more knowledge to answer a question.
After implicit intent is resolved by the reasoning question prompts, we can take full advantage of its knowledge to answer questions correctly.

\ignore{
\begin{table}
  \caption{Performance analysis of two steps.} 
  \label{tab:step}
  \begin{tabular}{l | c c c}
   \toprule
    Dataset & $k$ & Generation Step & Choose Step \\
    \midrule 
    OK-VQA & $5$ & $23.4$ & $20.3$\\
    test & $3$ & $22.3$ & $19.5$\\
    & $1$ & $19.4$ & $17.9$\\
    & $0$ & $-$ & $17.7$ \\
    \midrule
    A-OKVQA & $5$ & $33.7$ & $29.0$\\
    val & $3$ & $32.8$ & $28.3$\\
    & $1$ & $30.2$ & $27.2$ \\
    & $0$ & $-$ & $23.8$ \\
    \midrule 
    OK-VQA & $5$ & $23.4$ & $20.3$\\
    test & $3$ & $22.3$ & $19.5$\\
    & $1$ & $19.4$ & $17.9$\\
    & $0$ & $-$ & $17.7$ \\
  \bottomrule
\end{tabular}
\end{table}
}

\subsubsection*{\textbf{Ablation Study}}

We further evaluate the performance on different prompt design strategies and the results are displayed in Table~\ref{tab:step}.
If we eliminate the two-stage prompting, and simply choose the answer with highest $P(A)$ in Equation~(\ref{eq:pa}) as the final answer.
We have around $1$ point drop.
This indicates that the step of answer choosing is needed. It provides LLMs a chance to review the original question with the consideration of candidate answers.
During prompting for answer generation, we omit the aspects of scoring function in turn, the results indicate that all the aspects are important for generating a reasoning question prompt.
Among them, syntactic invariance is most significant aspect which measures the consistency of the substitution segments.
A replaced constituent with a different syntactic tagging easily leads to a chaotic sentence that cannot be understood by LLMs.
LM score and semantic integrity are also helpful in terms of measuring the sentence fluency and semantic integrity of the sentences.
During prompting for answer choosing, we omit the candidate construction and simply include the candidate answer without their confidence scores, which results in a performance drop.
After changing the confidence score to $P_{LLM}(A|\tilde{Q})$, the performance decreases, which indicates the importance of our answer heuristics construction.

\begin{table}
  \caption{Performance on A-OKVQA validation set having Img2Prompt as baselines but with different prompt designs.
  } 
  \label{tab:step}
  \begin{tabular}{l | c}
   \toprule
    Methods & A-OKVQA val \\
    \midrule 
    Img2Prompt+QR  prompt$_\text{\{GPT-3 175B\}}$ & $43.2$ \\
    \midrule
    w/o Two-stage prompting & $42.2$\\
    \midrule
    \multicolumn{2}{c}{\textit{Prompting for Answer Generation}} \\
    w/o LM score & $40.5$ \\
    w/o Semantic integrity & $40.8$\\
    w/o Syntactic invariance & $40.1$\\
    \midrule
    \multicolumn{2}{c}{\textit{Prompting for Answer Choosing}} \\
    w Plain answer heuristics & $42.0$ \\
    w/o Candidate construction & $42.8$ \\
  \bottomrule
\end{tabular}
\end{table}

\subsubsection*{\textbf{Case Study}}

We display some cases in Figure~\ref{fig:case} to investigate how our reasoning question prompts work in zero-shot VQA tasks.

Example (a) contains an image of a blow dryer, the generated caption is ``\textit{This is a blow dryer in a bathroom}''.
The visual question is ``\textit{What is the appliance the woman is holding used for?}''.
As we can see, this is an ill-posed question as there is no woman shown on the image.
The result of LLMs without any reasoning question prompt is ``\textit{cutting hair}'', which may caused by the unexpected bias of the LLMs.
Based on the caption and question, we generate reasoning question prompts such as ``\textit{What is the appliance a blow dryer used for?}'' and ``\textit{What is the appliance a bathroom is holding used for?}'', which successfully bridge the gap between the image and the question, thus LLMs can predict correct answer.
Similarly, in example (b), there is a gap between ``\textit{the child eating}'' and the image.
The queried objective is not explicitly mentioned in the question, so LLMs must infer the object that the question is asking about.
Reasoning question prompts such as ``\textit{What is dishes in front of her?}'' and ``\textit{What is a cup with food in dishes in front of her?}'' explicate the queried object so that LLMs can easily understand the question and return the correct answers.
More cases can be found in Appendix~\ref{ap:case}.

\section{Conclusion}
\label{sec:con}

In this paper, we investigate zero-shot VQA tasks via LLMs, where images are first converted into captions then LLMs answer questions based on the caption contents.
We propose a way to generate reasoning question prompts, which can help explicate the intermediate reasoning step of a question and eliminate the semantic gap between the question and the caption.
The experiments show that reasoning question prompts improve existing zero-shot VQA methods with different LLM backbones and achieve a new state-of-the-art performance on multiple zero-shot VQA data sets.

\section*{Acknowledgements}

The authors would like to thank the anonymous reviewers for
their insightful comments. 
This work was supported by the Natural Science Foundation of China (Project No. 62206097) and Shanghai Pujiang Talent Program (Project No. 22PJ1403000).

\bibliographystyle{ACM-Reference-Format}
\balance
\bibliography{software}

\appendix

\newpage

\section{Appendix}

\subsection{Details about Unsupervised Question Edition}
\label{ap:edition}

For each sentence, we use CoreNLP\footnote{\url{https://stanfordnlp.github.io/CoreNLP/}} to construct the constituency tree and Spacy\footnote{\url{https://spacy.io/}} to obtain the part-of-speech and dependency tags of the words.

The syntax-aware LM model we used takes words, POS tags and dependency tags as the input, which can be denoted as:
\begin{equation*}
 \mathbf{w} = [\mathbf{v}(w); \mathbf{p}(w); \mathbf{d}(w)],
\end{equation*}
where $\mathbf{v}(w)$ is the word embeding, $\mathbf{p}(w)$ is the POS tag embedding and $\mathbf{d}(w)$ is the dependency tag embedding.
The dimensions of POS tag and dependency tag are $150$, and the dimension of word embedding is $300$.
$\mathbf{w}$ is fed into the LM~\cite{kumar:acl2020}, which enables a LM to be sensitive to the sentence structure.
We directly take the checkpoint\footnote{\url{https://github.com/ddhruvkr/Edit-Unsup-TS}} of the syntax-aware LM model in prior study~\cite{kumar:acl2020} on text simplification as the initialization.
This checkpoint is initially trained on WikiLarge datasets.
We fine-tune the model with questions in OK-VQA test data, so that the LM model can be quickly adapted to VQA domains.
Regarding fine-tuning, we use the Stochastic Gradient Descent algorithm with $0.4$ as the dropout rate and $32$ as the batch size.

\subsection{Details of Re-implementation of Img2Prompt on VQAv2 Dataset}
\label{ap:img2prompt}

We re-implement the source code of Img2Prompt to generate synthetic examples.
Due to the large size of the VQAv2 dataset and our limited computational resource, we implement a light version.
Specifically, Img2prompt leverages BLIP to generate captions from a given image and conduct image-question matching.
In the official implementation setting, they sample $10$ image patches and then generate $100$ question-relevant captions, from which they can produce $30$ question-answer pairs.
For us, we sample $10$ image patches for each image but simply generate $20$ captions by adjusting the number of the generation.
Based on these $20$ captions, we subsequently generate $10$ question-answer pairs.
These question-answer pairs are utilized as the exemplar prompts for answer generation and answer choosing.
Therefore, there might be some information loss in our implementation as some important exemplars might be filtered out.

\subsection{RQ Prompts with Different LLMs}
\label{ap:diff_llm}

\begin{table}
  \caption{Zero-shot performance A-OKVQA validation set having Img2Prompt as baselines but with different LLMs.
  $\Delta$ denotes the performance gain brought by QR prompts.} 
  \label{tab:llm}
  \begin{tabular}{l c c c}
   \toprule
    LLMs & Img2Prompt & +QR prompt & $\Delta$ \\
    \midrule 
    GPT-3 175B & $38.9$ & $43.2$ & $\uparrow 4.3$\\
    \midrule
    GPT-3.5 175B & $37.1$ & $40.3$ & $\uparrow 3.2$ \\
    GPT-Neo 2.7B & $29.7$  & $31.5$ & $\uparrow 1.8$\\
    BLOOM 7.1B & $29.8$ & $32.1$ & $\uparrow 2.3$\\
    GPT-J 6B & $32.5$& $33.1$ & $\uparrow 0.6$\\
    OPT 125M & $10.8$ & $13.3$ & $\uparrow 2.5$\\
  \bottomrule
\end{tabular}
\end{table}

To verify the scaling effect of reasoning question prompts with different LLMs, we conduct experiments on A-OKVQA validation set having Img2Prompt as baselines but with different LLMs.
The result is displayed in Table~\ref{tab:llm}.
Specifically, we evaluate on GPT-3.5 $175B$, GPT-Neo $2.7$B~\cite{Black:Zenodo2021}, BLOOM $7.1$B~\cite{scao:arXiv2022}, GPT-J $6$B~\cite{Wang:github2021} and OPT-$125$M~\cite{zhang:arxiv2022}.
As we can see, the performance of zeros-shot VQA tasks is affected by the size of LLMs.
A LLM with larger model size usually results in a better performance, which is also verified in prior paper~\cite{guo:cvpr2023}.
Importantly, including reasoning question prompts can always improve the performance, which further verifies the generalization capability of our method.

\subsection{Case Study}
\label{ap:case}

We display more examples in OK-VQA test set to show how the reasoning question prompts work in zero-shot VQA tasks.
The displayed examples are predicted by Img2Prompt+QR prompt$_\text{ \{OPT-30B\}}$ and the original predictions without QR prompts are incorrect.
In the Figure, the edited segments of the question are highlighted with red color and the correct predicted answers are highlighted with green color.

\begin{minipage}{.5\textwidth}
  \centering
  \captionsetup{labelformat=empty} 
  \includegraphics[width=\linewidth]{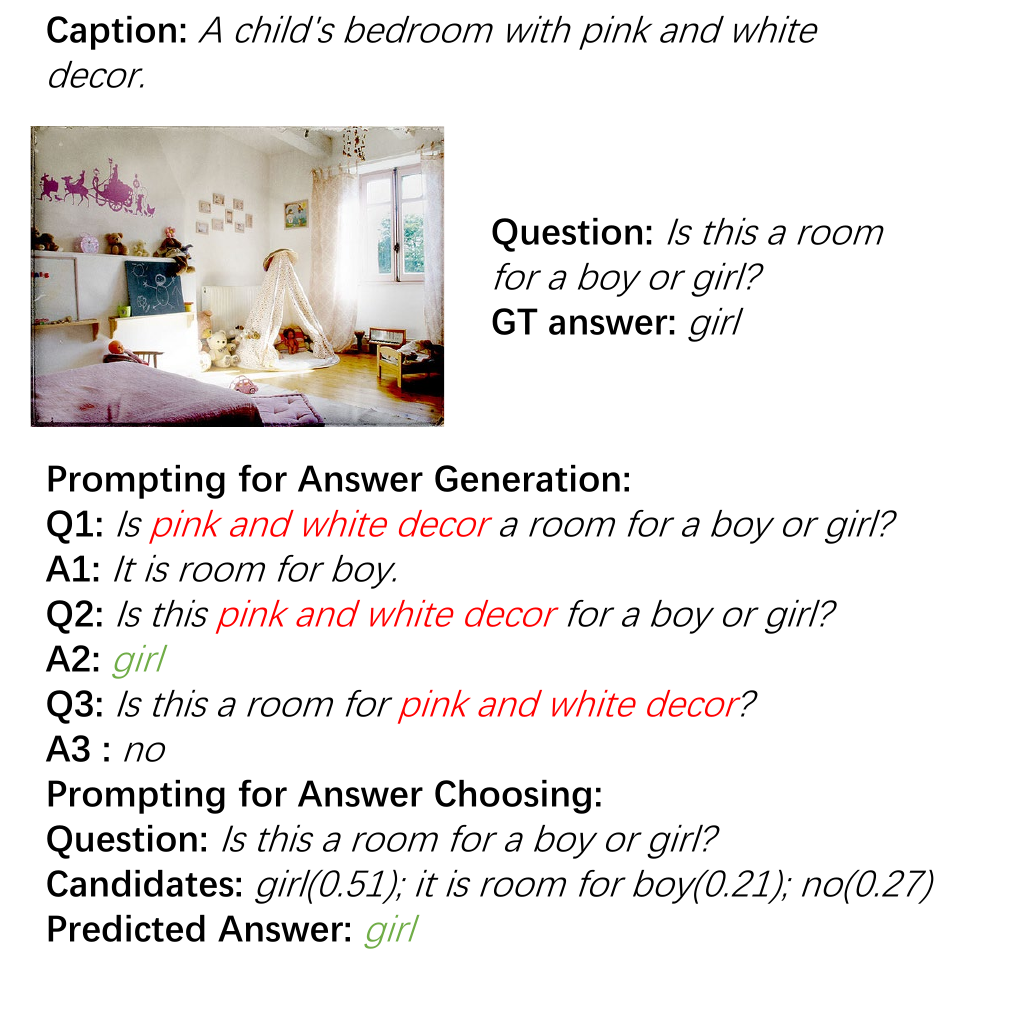}
\vspace{-0.8cm}
  \captionof{figure}{Example 1}
\vspace{1cm}
  \label{fig:test7}
\end{minipage}

\begin{minipage}{.5\textwidth}
  \centering
  \captionsetup{labelformat=empty} 
  \includegraphics[width=\linewidth]{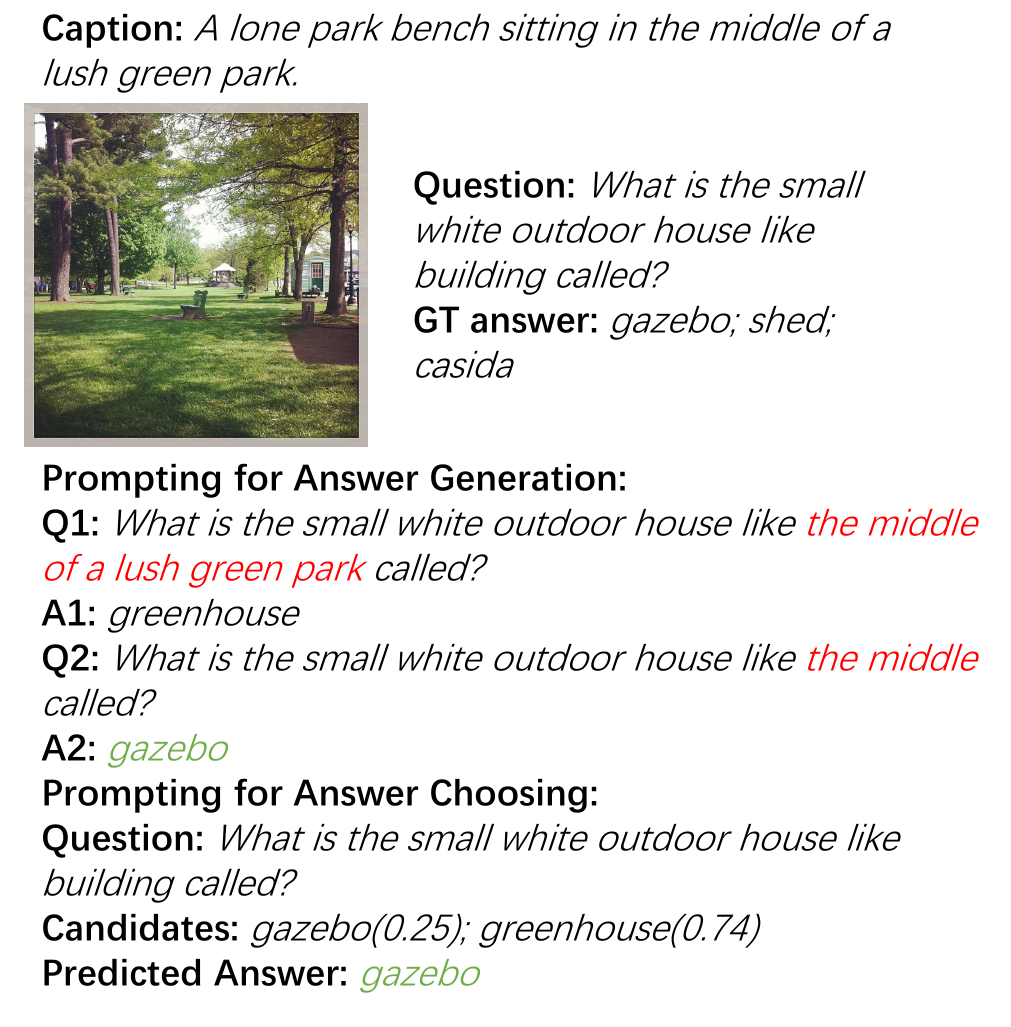}
\vspace{-0.8cm}
  \captionof{figure}{Example 2}
\vspace{1cm}
  \label{fig:test8}
\end{minipage}

\begin{minipage}{.5\textwidth}
  \centering
  \captionsetup{labelformat=empty} 
  \includegraphics[width=\linewidth]{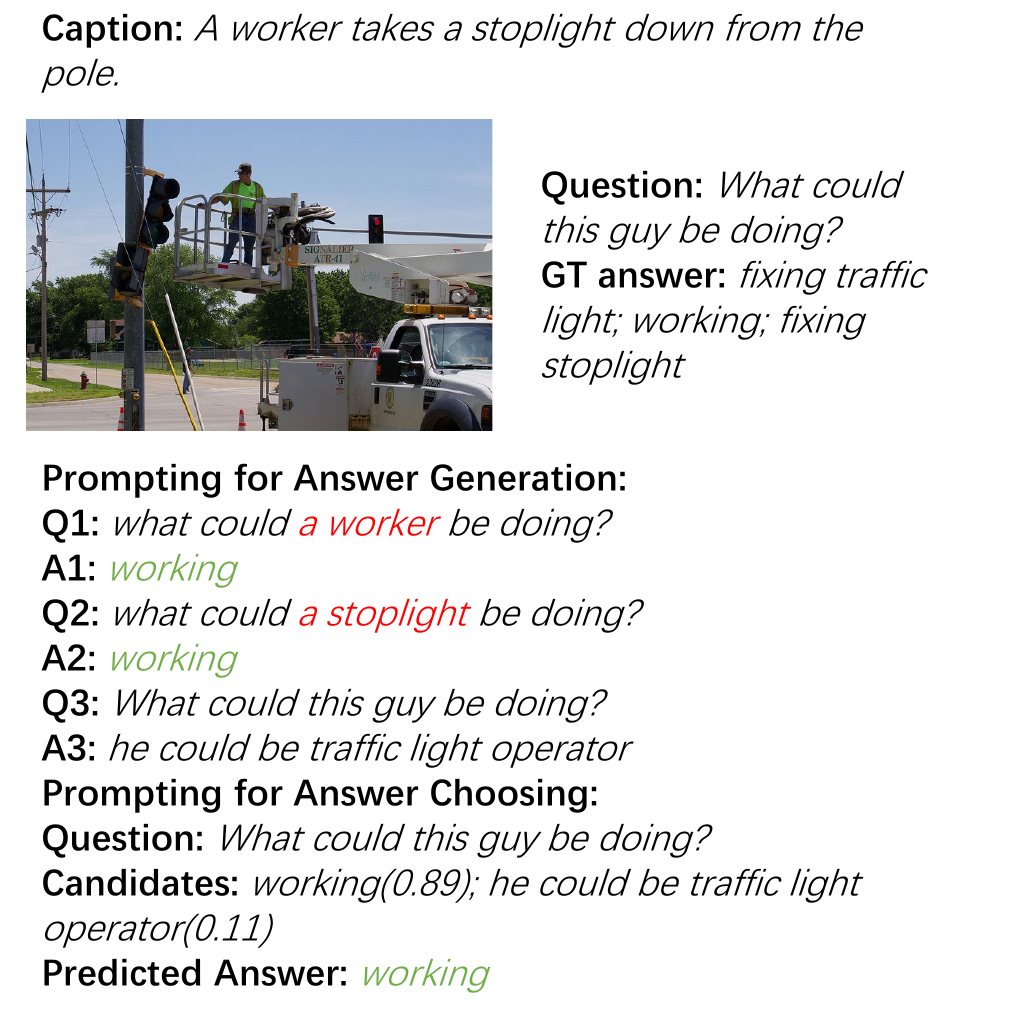}
\vspace{-0.8cm}
  \captionof{figure}{Example 3}
\vspace{1cm}
  \label{fig:test9}
\end{minipage}

\begin{minipage}{.5\textwidth}
  \centering
  \captionsetup{labelformat=empty} 
  \includegraphics[width=\linewidth]{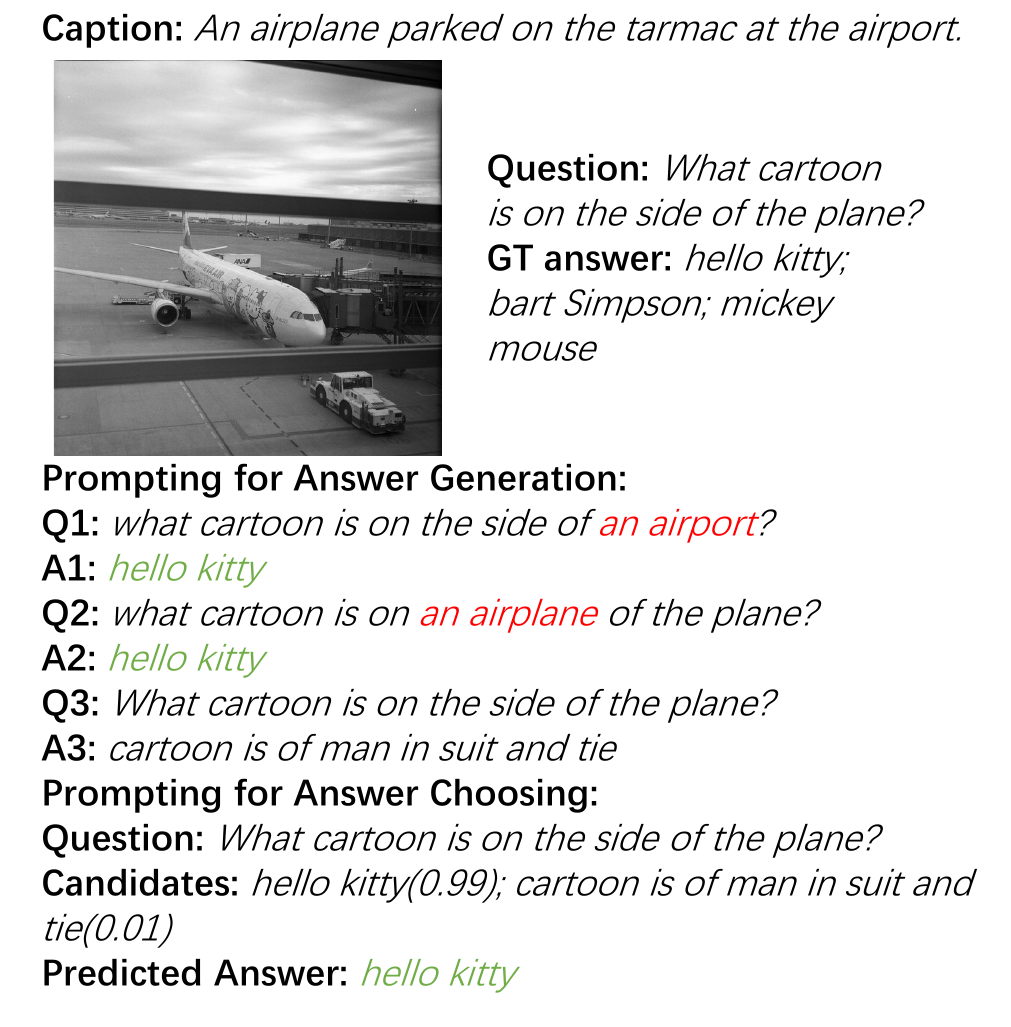}
\vspace{-0.8cm}
  \captionof{figure}{Example 4}
\vspace{1cm}
  \label{fig:test10}
\end{minipage}

\begin{minipage}{.5\textwidth}
  \centering
  \captionsetup{labelformat=empty} 
  \includegraphics[width=\linewidth]{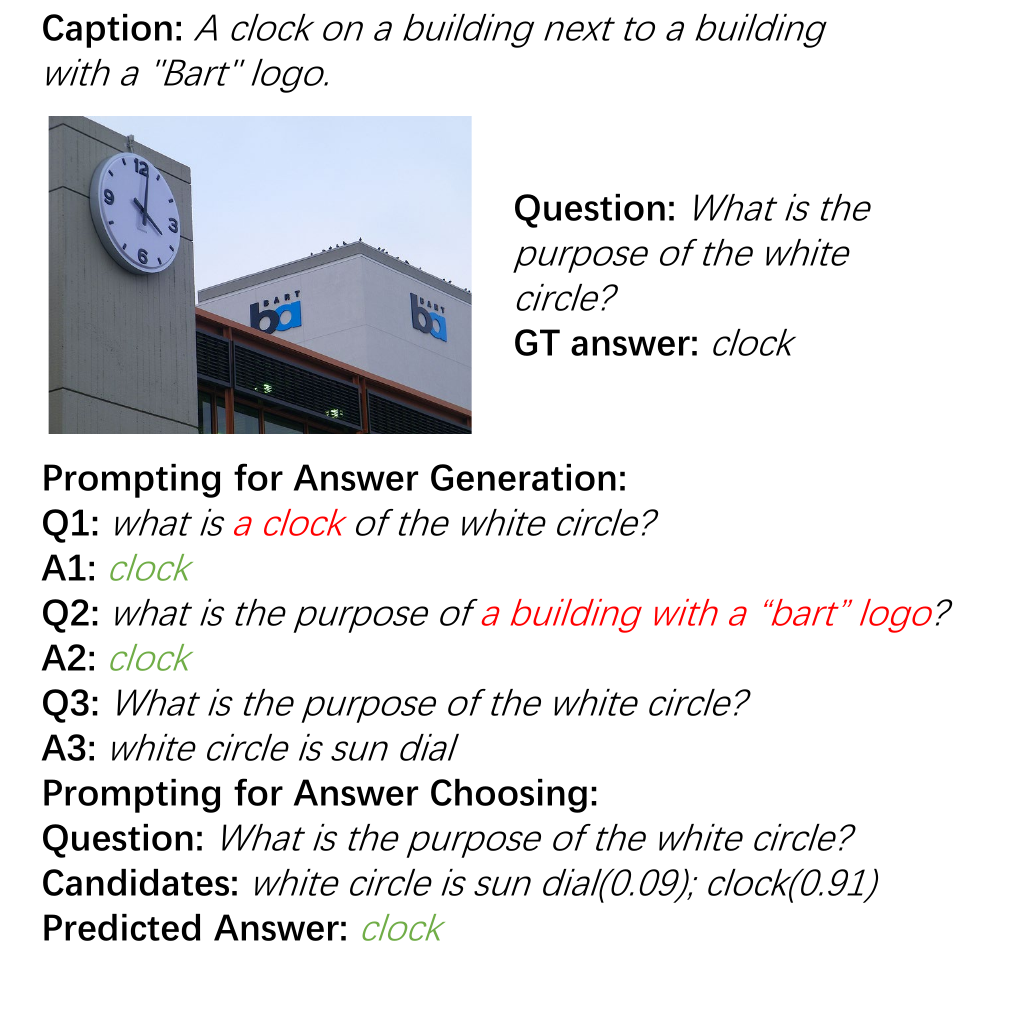}
\vspace{-0.8cm}
  \captionof{figure}{Example 5}
\vspace{1cm}
  \label{fig:test11}
\end{minipage}




\end{document}